\documentclass[runningheads]{llncs}
\usepackage{graphicx}

\usepackage{cite}
\usepackage{amsmath,amssymb,amsfonts}
\usepackage{algorithmic}
\usepackage{textcomp}
\usepackage{xcolor}

\usepackage{caption}
\usepackage{bbm}
\usepackage{subfigure}
\usepackage{booktabs}
\usepackage{multirow}

\usepackage{marvosym}
\begin{document}
\title{SiamSNN: Siamese Spiking Neural Networks for Energy-Efficient Object Tracking\thanks{Supported in part by the National Natural Science Foundation of China under Grant 61572214, 62006070, and Seed Foundation of Huazhong University of Science and Technology (2020kfyXGYJ114).}}
%
%
\author{Yihao Luo\inst{1} \and
	Min Xu\inst{1} \and
	Caihong Yuan\inst{1,2}  \and
	Xiang Cao \inst{1}  \and
	Liangqi Zhang \inst{1}  \and
	Yan Xu \inst{1}  \and
	Tianjiang Wang\inst{1} \and
	Qi Feng\inst{1}\textsuperscript{(\Letter)}
}
\authorrunning{Y. Luo et al.}
%
\institute{
	Huazhong University of Science and Technology, Wuhan 430074, China\\
	\email{fengqi@hust.edu.cn}\\
	Henan University, Kaifeng 475004, China
}
\maketitle              
\begin{abstract}
Recently spiking neural networks (SNNs), the third-generation of neural networks has shown remarkable capabilities of energy-efficient computing, which is a promising alternative for deep neural networks (DNNs) with high energy consumption.
SNNs have reached competitive results compared to DNNs in relatively simple tasks and small datasets such as image classification and MNIST/CIFAR, while few studies on more challenging vision tasks on complex datasets.
In this paper, we focus on extending deep SNNs to object tracking, a more advanced vision task with embedded applications and energy-saving requirements,
and present a spike-based Siamese network called SiamSNN.
Specifically, we propose an optimized hybrid similarity estimation method to exploit temporal information in the SNNs, and introduce a novel two-status coding scheme to optimize the temporal distribution of output spike trains for further improvements.
SiamSNN is the first deep SNN tracker that achieves short latency and low precision loss on the visual object tracking benchmarks OTB2013/2015, VOT2016/2018, and GOT-10k.
Moreover, SiamSNN achieves notably low energy consumption and real-time on Neuromorphic chip TrueNorth.

\end{abstract}

\section{Introduction}
\label{sec:introduction}
Nowadays Deep Neural Networks (DNNs) have shown remarkable performance in various scenarios~\cite{yuan2019learning,guo2020object,CE-FPN}. However, it is hard to employ DNNs on embedded systems such as mobile devices due to the high computation cost and heavy energy consumption.
Spiking neural networks (SNNs) are regarded as the third generation artificial neural networks,
which are biologically-inspired computational models and have realized ultra-low power consumption on neuromorphic hardware (such as TrueNorth~\cite{merolla2014million}).
But they still lag behind DNNs in terms of accuracy~\cite{tavanaei2018deep}. One major reason is that the training algorithms can only train shallow SNNs~\cite{park2019fast}.
To avoid this issue, an alternative way is converting trained DNNs to SNNs~\cite{CaoSpiking,diehl2015fast,rueckauer2017conversion},
which aims to convert DNNs to SNNs by transferring the well-trained models with weight rescaling and normalization methods.

Although SNNs have shown excellent energy-efficient potential, they have been limited to relatively simple tasks and small datasets (e.g., image classification on MNIST and CIFAR)~\cite{wu2019direct}.
Most critically, the insufficiency of spike-based modules for various functions outside classification limits their applications~\cite{kim2020spiking}, it is necessary to enrich them for greater prevalence. 
Besides, SNNs of remarkable performance through conversion usually requires long latency~\cite{rueckauer2017conversion,sengupta2019going}, which causes the inability to reach real-time requirements. Exploiting the sparse and temporal dynamics of spike-based information is significant to achieve real-time algorithms~\cite{roy2019towards}.

Following the successes of the DNNs to SNNs conversion methods on image classification and object detection tasks~\cite{kim2020spiking}, it is desirable to extend those results to solve visual object tracking, a more advanced vision task with embedded applications and energy-saving requirements.
The state-of-the-art tracking models~\cite{li2019siamrpn++,XuWLYY20,Voigtlaender2020Siam} require high-performance GPUs, which involve a drastic increase in computation and power requirements that tends to be intractable for embedded platforms~\cite{basu2018low}.
With the aid of energy-efficient computing in SNNs, implementing a spike-based tracker is significant to realize energy-efficient tracking algorithms on embedded systems, which also extends the limitations of related theories and applications of SNNs. The problems of implementing it can be summarized as follows: 1) There is no effective similarity estimation function in SNN fashion for deep tracking. 2) The existed methods usually require long latency while real-time is a significant metric to tracking tasks.

In this paper, we focus on extending deep SNNs to object tracking with DNN-to-SNN conversion methods. To address the first issues, we introduce an optimized hybrid similarity estimation method in SNN fashion. The proposed method evaluates the similarity between feature maps of the exemplar image and candidate images in the form of spiking. And inspired by the neural phenomenon of long-term depression (LTD)~\cite{bi1998synaptic}, we design a two-status coding scheme to optimize the temporal distribution of output spike trains, which utilizes temporal information for shortening the latency to reach real-time and mitigating the accuracy degradation. 
Eventually, we present a spike-based Siamese network for object tracking called SiamSNN based on SiamFC~\cite{bertinetto2016fully}. To the best of our knowledge, SiamSNN is the first deep SNN for object tracking that achieves short latency and low precision loss of the original SiamFC on the tracking benchmarks VOT2016~\cite{vot2016}, VOT2018~\cite{vot2018}, and GOT-10k~\cite{huang2019got}.

\section{Related Work and Background}

\subsection{Conversion methods of DNN-to-SNN}

Converting trained DNNs to SNNs is an effective approach to obtain deep SNNs with relatively high accuracy.
Cao $\textit{et al.}$~\cite{CaoSpiking} convert DNNs to SNNs and achieve excellent performance with a comprehensive conversion scheme. Then Rueckauer $\textit{et al.}$~\cite{rueckauer2017conversion} introduce spike max-pooling and batch normalization (BN) for SNN.
Different from DNNs, SNNs use event-driven binary spike trains in a certain period rather than a single continuous value between neurons. The widely used integrate-and-fire (IF) neuron model integrates post-synaptic potential (PSP) into the membrane potential \begin{math}V_\mathrm{mem}\end{math}. In other words, the spiking neuron \begin{math}i\end{math} accumulates the input \begin{math}z\end{math} into the membrane potential \begin{math}V_{\mathrm{mem}, i}^{l}\end{math} in the \begin{math}l\end{math}th layer at each time step, which is described as
\begin{equation}
V_{\mathrm{mem}, i}^{l}(t)=V_{\mathrm{mem}, i}^{l}(t-1)+z_{i}^{l}(t)-V_{\mathrm{th}}(t)\Theta_{i}^{l}(t),
\label{eq1}
\end{equation}
where \begin{math}\Theta_{i}^{l}(t)\end{math} is a spike in the \begin{math}i\end{math}th neuron in the \begin{math}l\end{math}th layer, \begin{math}V_{\mathrm{th}}(t)\end{math} is a certain threshold, and \begin{math}z_{i}^{l}(t)\end{math} is the input of the \begin{math}i\end{math}th neuron in the \begin{math}l\end{math}th layer, which is described as
\begin{equation}
z_{i}^{l}(t)=\sum_{j} w_{j i}^{l} V_{\mathrm{th}}(t)\Theta_{j}^{l-1}(t)+b_{i}^{l},
\label{eq2}
\end{equation}
where \begin{math}w^{l}\end{math} is the weight and \begin{math}b^{l}\end{math} is the bias in the \begin{math}l\end{math}th layer. Each neuron will generate a spike only if its membrane potential exceeds a certain threshold \begin{math}V_{\mathrm{th}}(t)\end{math}. The process of spike generation can be generalized as
\begin{equation}
\Theta_{i}^{l}(t)=U\left(V_{\mathrm{mem}, i}^{l}(t-1)+z_{i}^{l}(t)-V_{\mathrm{th}}(t)\right),
\label{eq3}
\end{equation}
where \begin{math}U\left(\cdot\right)\end{math} is a unit step function. Once the spike is generated, the membrane potential is reset to the resting potential. As shown in Eq.~\ref{eq1}, we adopt the method of resetting by subtraction rather than resetting to zero, which increases firing rate and reduces the loss of information. The firing rate is defined as
\begin{equation}
\text { firing rate }=\frac{N}{T},
\label{eqFR}
\end{equation}
where N is the total number of spikes during a given period \begin{math}T\end{math}. The maximum firing rate will be 100\% since the neuron generates a spike at each time step.

To prevent the neuron from over-activation and under-activation, Diehl $\textit{et al.}$~\cite{diehl2015fast} suggest a data-based method to enable a sufficient firing rate and achieve nearly lossless accuracy on MNIST.
Sengupta $\textit{et al.}$~\cite{sengupta2019going} introduce a novel spike-normalization method for generating an SNN with deeper architecture and experiment on complex standard vision dataset ImageNet~\cite{russakovsky2015imagenet}.
The conventional normalized weights \begin{math}{w}^{l}\end{math} and biases \begin{math}{b}^{l}\end{math} are calculated by
\begin{equation}
\tilde{w}^{l}=w^{l} \frac{\lambda^{l-1}}{\lambda^{l}} \quad \text { and } \quad \tilde{b}^{l}=\frac{b^{l}}{\lambda^{l}},
\label{eqlnorm}
\end{equation}
where \begin{math}w\end{math} are the weights of the original DNN, \begin{math}b\end{math} are the biases, and \begin{math}\lambda\end{math} is the 99.9th percentile of the maximum activation~\cite{rueckauer2017conversion} in the layer \begin{math}l\end{math}.
And Spiking-YOLO~\cite{kim2020spiking} proposes a channel normalization for SNN object detection.

As for spike coding, the method of representing information with spike trains, rate and temporal coding are the most frequently used coding schemes in SNNs.
Rate coding is based on the spike firing rate, which is used as the signal intensity, counted by the number of spikes that occurred during a period. Thus, it has been widely used in converted SNNs~\cite{CaoSpiking,rueckauer2017conversion} for classification by comparing the firing rates of each category. However, rate coding requires a higher latency for more accurate information.
And temporal coding uses timing information to mitigate long latency. 
Park $\textit{et al.}$~\cite{park2019fast} introduce a burst coding scheme inspired by neuroscience research and investigates a hybrid coding scheme to exploit different features of different layers in SNNs.
Kim $\textit{et al.}$~\cite{kim2018deep} propose weighted spikes by the phase function to make the most of the temporal information. 
The phase function of it is given by
\begin{equation}
V_{\mathrm{th}}(t)=2^{-(1+\bmod (t-1, K))}
\label{eqphase}
\end{equation}
where K is the period of the phase. Phase coding needs only K time steps to represent K-bit data (8-bit images usually) whereas rate coding would require \begin{math}2^K\end{math} time steps to represent the same data.

\subsection{Object tracking}
\label{sectrack}

Object tracking aims to estimate the position of an arbitrary target in a video sequence while its location only initializes in the first frame by a bounding box. 
Two main branches of recent trackers are based on correlation filter (CF)~\cite{danelljan2017eco} and DNNs~\cite{li2018high}. CF trackers train regressors in the Fourier domain and update the weights of filters to do online tracking. Motivated by CF, trackers based on Siamese networks with similarity comparison strategy have drawn great attention due to their high performance. SiamFC~\cite{bertinetto2016fully} introduces the correlation layer rather than fully connected layers to evaluate the regional feature similarity between two frames.
It learns a function \begin{math}f(z, x)\end{math} to compare an exemplar image \begin{math}z\end{math} to candidate images \begin{math}x\end{math} and computes the similarity at all sub-windows on a dense grid in a single evaluation, which is defined as
\begin{equation}
f(z, x)=\varphi(z) * \varphi(x)+b\cdot\mathbbm{1},
\label{eqSiamFC}
\end{equation}
where \begin{math}\varphi\end{math} is the CNN to perform feature extraction and \begin{math}b\cdot\mathbbm{1}\end{math} denotes a bias equated in every location.
In the subsequent work, SiamRPN~\cite{li2018high}, SiamRPN++~\cite{li2019siamrpn++}, Siam R-CNN~\cite{Voigtlaender2020Siam}, SiamFC++~\cite{XuWLYY20} obtain the state-of-the-art tracking performance.
All of them are based on the Siamese architecture in SiamFC.
Therefore, we select SiamFC as our base model to construct a Siamese Spiking Neural Network for energy-efficient object tracking.

For the aspect of energy and memory consumption, Liu $\textit{et al.}$~\cite{liu2019teacher} construct small, fast yet accurate trackers by a teacher-students knowledge distillation model.
To exploit the low-powered nature of SNNs, Yang $\textit{et al.}$~\cite{yang2019dashnet} propose a hybrid paradigm of ANN and SNN for efficient high-speed object tracking, called Dashnet.
However, Dashnet is not an absolute deep SNN, the problem of insufficient computing power still exists while applying to resource-constrained systems.
Luo $\textit{et al.}$~\cite{luo2019spiking} show the possibility to construct a deep SNN for object tracking (the early version of SiamSNN) on some simple videos.
But they lack effective methods and convincing performance on the benchmark.
We propose two approaches to obtain a stronger spiking tracker in this work.

\begin{figure*}[!t]
	\centering
	\includegraphics[width=0.8\linewidth]{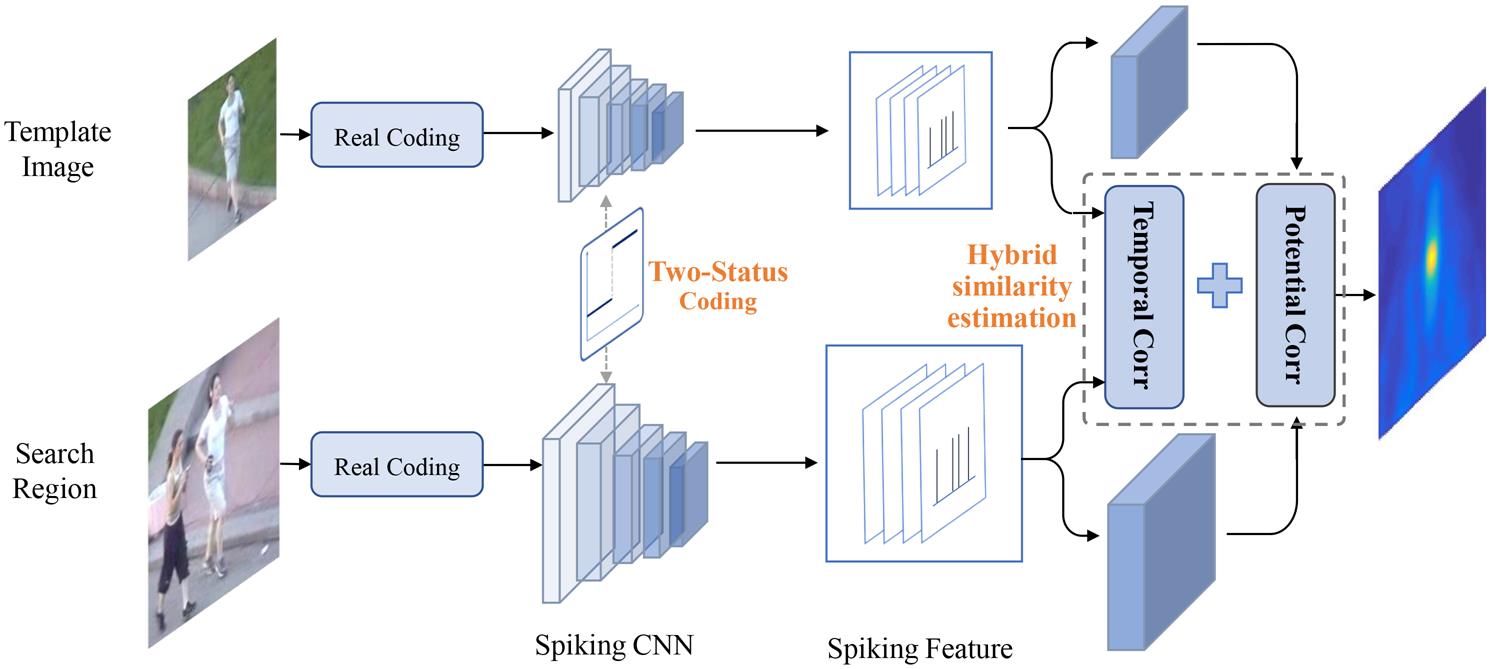}
	\caption{\textbf{Framework of SiamSNN}. This framework consists of a converted Siamese spiking convolutional neural network, two-status coding scheme, and an optimized hybrid similarity estimation method (potential and temporal correlation).}
	\label{figSNN}
\end{figure*}

\section{Proposed Methods}

\subsection{Model overview}

We detail the SiamSNN, the converted SiamFC in Fig.~\ref{figSNN}. SiamSNN consists of a converted Siamese spiking convolutional neural network, two-status coding scheme, and an optimized hybrid similarity estimation method (potential and temporal correlation).
The Spiking CNN is consistent with the CNN model architecture in SiamFC~\cite{bertinetto2016fully},
which can be obtained by Eq.~\ref{eq1},\ref{eq2},\ref{eq3},\ref{eqlnorm}.
The hybrid similarity estimation method calculates the response map with the potential and temporal correlation of spike features.
The two-status coding scheme optimizes the temporal distribution of output spike trains to improve the performance. 

Similar to spiking softmax layer~\cite{rueckauer2017conversion}, Eq.~\ref{eqSiamFC} can be converted into a spiking form:
\begin{equation}
M_P(z, x)=\left(\sum_{t=1}^T\varphi(z(t)) \right)* \left(\sum_{t=1}^T\varphi(x(t)) \right),
\label{eqV}
\end{equation}
where \begin{math}M_P\end{math} is the potential response map, \begin{math}z(t)\end{math} and \begin{math}x(t)\end{math} are the input encoded spike trains of exemplar image and candidate image, \begin{math}\varphi\end{math} is the Spiking CNN, \begin{math}T\end{math} is the latency.
We will refer to this simple method as potential similarity estimation (PSE) in the rest of the paper.

\begin{figure}[!t]
	\centering
	\subfigure[Temporal correlation between spike train A and B.]{
		\label{figtrains:a} 
		\includegraphics[width=0.45\linewidth]{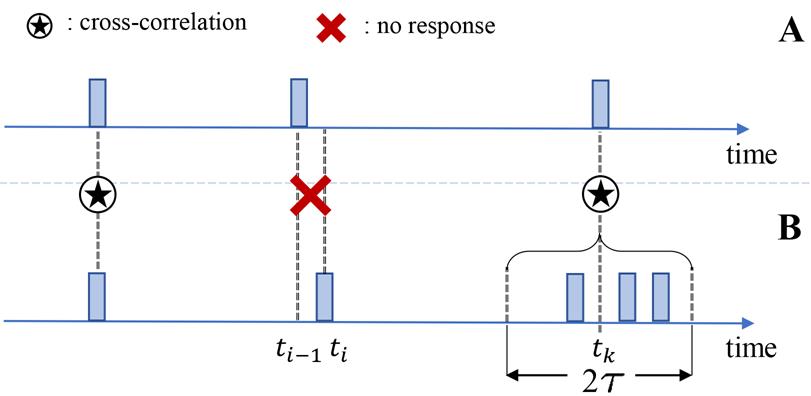}}
	\subfigure[The distribution of spikes after two-status coding.]{
		\label{figtrains:b} 
		\includegraphics[width=0.45\linewidth]{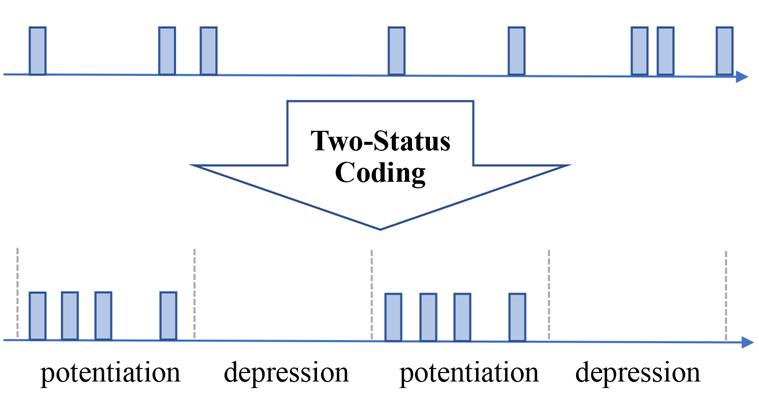}}
	\caption{Illustrations of the hybrid spiking similarity estimation and two-status coding.}
\end{figure}

\subsection{Hybrid spiking similarity estimation}

PSE does not consider the temporal information of spike trains, which is the main difference between DNNs and SNNs and significant to information transmission~\cite{park2019fast}.
We can also match the temporal similarity between spiking features of exemplar and candidate images as
\begin{equation}
M_T(z, x)=\sum_{t=1}^T\left(\varphi(z(t))* \varphi(x(t)) \right),
\label{eqT}
\end{equation}
which is similar to Eq.~\ref{eqV}, \begin{math}M_T\end{math} is the temporal response map. It calculates the correlation at each time step.
We will refer to it as temporal similarity estimation (TSE) in the rest of the paper.

Intuitively, rely on the synchronization of spikes at each time step, TSE causes strict temporal consistency of correlation. For instance in Fig.~\ref{figtrains:a}, if spike train A fires at \begin{math}t_{i-1}\end{math}, and spike train B fires at \begin{math}t_i\end{math}, their temporal correlation equals 0 during \begin{math}(t_{i-2},t_{i+1})\end{math}, but they are highly similar measured by spike-distance. Motivated by PySpike~\cite{mulansky2016pyspike}, we optimize TSE and propose a response period with time weights during correlation operation. It is defined as:
\begin{equation}
M_{\Delta\tau}(z, x)=\sum_{t=1}^{T}\sum_{m=t-\tau}^{t+\tau}\left(\frac{\varphi(z(m))* \varphi(x(t)) }{2|t-m|+1}\right),
\label{eqtau}
\end{equation}
where \begin{math}\tau\end{math} is the response period threshold. We set \begin{math}\tau = 1\end{math} by experiments, and the impacts of \begin{math}\tau\end{math} are analyzed in Fig.~\ref{figlatencytao:b} of Section~\ref{sechp}. As shown in Fig.~\ref{figtrains:a}, Eq.~\ref{eqtau} makes spike trains A response with B during \begin{math}[t_k-\tau,t_k+\tau])\end{math}. The response value will attenuate gradually when the time steps far from \begin{math}t_k\end{math}.
A large period has few contributions to the response value and also increases computation consumption.
Thus, \begin{math}\tau\end{math} is usually set to a small number.

Considering PSE performs lossless on portions of sequences in OTB-2013, we calculate the final response map as follows:
\begin{equation}
f_{spike}(z, x) = \frac{1}{T^2}M_P(z, x) + \frac{1}{2T\cdot2\tau}M_{\Delta\tau}(z, x)+b\cdot\mathbbm{1},
\label{eqres}
\end{equation}
where the weights of \begin{math}M_P\end{math} and \begin{math}M_{\Delta\tau}\end{math} aim to normalize them to the same magnitude, \begin{math}b\cdot\mathbbm{1}\end{math} denotes the same bias in SiamFC. Noted that in SiamFC, \begin{math}f(z,x)\end{math} compares an exemplar image \begin{math}z\end{math} to a candidate image \begin{math}x\end{math} in the current frame of the same size. It generates a high value if the two images describe the same object and low otherwise.
And \begin{math}f_{spike}(z, x)\end{math} is the SNN fashion of this function, not strictly calculates the similarity between two spike trains (e.g., the similarity equals 1 but response equals 0 between two zero spike trains).
It is why choosing \begin{math}T\end{math} instead of spike numbers as the normalization parameter to get the response value.
The proposed similarity estimation method makes it possible to convert the state-of-the-art Siamese trackers into deep SNNs.
In the rest of the paper, we will refer to hybrid spiking similarity estimation as HSE.

\subsection{Two-status coding scheme}
\label{sects}

Although Eq.~\ref{eqres} enhances the temporal correlation between two spike trains, the temporal distributions of spikes are random and disorganized, which makes a large portion of spike values underutilized.
For shortening latency and reducing energy consumption, we expect that the output spike of each time step will contribute to the final response value.

Phase coding (Eq.~\ref{eqphase}) indicates that assigning periodicity to spike trains enhances their ability of information transmission, which brings efficient performance in image classification.
Therefore, we draw on the experience of periodic spiking feature when estimating the similarity.
According to~\cite{bi1998synaptic}, repetitive electrical activity can induce a persistent potentiation or depression of synaptic efficacy in various parts of the nervous system (LTP and LTD).
Motivated by this neural phenomenon, we propose two-status coding scheme. It represents the voltage threshold of potentiation status and depression status. We define the function as follows:
\begin{equation}
V_{\mathrm{th}}(t)=\left\{\begin{array}{ll}{\alpha} & {\text {if} \bmod(t/p,2) = 1 } \\ {\beta} & {\text { otherwise , }}\end{array}\right.
\end{equation}
where \begin{math}\alpha \to+\infty\end{math} to prevent neurons from generating spikes during depression status and \begin{math}\beta\end{math} is often smaller than the normalized voltage threshold \begin{math}1\end{math} to excite neurons spiking during potentiation status, \begin{math}p\end{math} is a constant that controls the period of neuron state change, \begin{math}t\end{math} is the current time step. We set \begin{math}p = 5\end{math} by observation of experiments and  \begin{math}\beta = 0.5\end{math} by experiences in Section~\ref{sechp}.

Our two-status coding makes neurons fire with the same value in potentiation status and accumulates membrane potential in depression status.
It constrains equivalent spikes in a fixed periodic distribution, increase the density of spikes in potentiation status to enhance response value.
Fig.~\ref{figtrains:b} shows the distribution of spikes after two-status coding from the original rate coding. The proposed method can also save energy by avoiding neurons generating spikes which are useless for HSE. Moreover, spiking neurons in the next layer are not consuming energy during depression status due to zero input.

Park $\textit{et al.}$~\cite{park2019fast} propose a hybrid coding scheme by the motivation that neurons occasionally use different neural coding schemes depending on their roles and locations in the brain. Inspired by this idea, we use two-status coding scheme to optimize the temporal distribution of output spikes, and real coding for the input layer due to its fast and accurate features.

\section{Experiments}

\subsection{Datasets and metrics}

We evaluate our methods on OTB-2013~\cite{wu2013online}, OTB-2015~\cite{wu2015object}, VOT2016~\cite{vot2016}, VOT2018~\cite{vot2018}, and GOT-10k~\cite{huang2019got}. The simulation and implementations are based on TensorFlow.
To assess and verify the accuracy of the proposed methods, we use the average overlap ratio as the basic metric.
As for energy consumption, we average the number of frames in a random video.

\begin{figure}[!t]
	\centering
	\subfigure[ ]{
		\label{figlatencytao:a} 
		\includegraphics[width=0.45\linewidth]{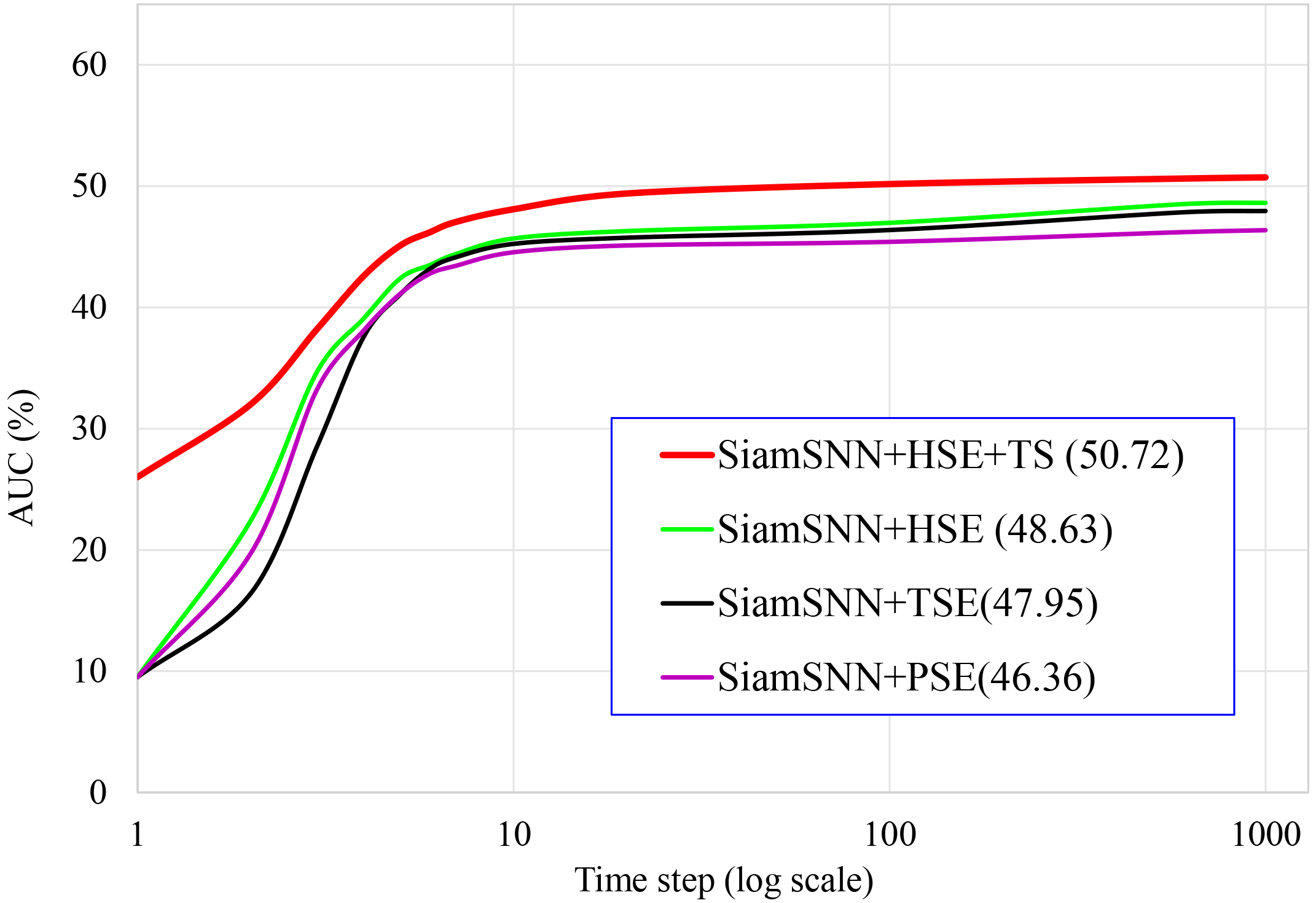}}
	\subfigure[ ]{
		\label{figlatencytao:b} 
		\includegraphics[width=0.45\linewidth]{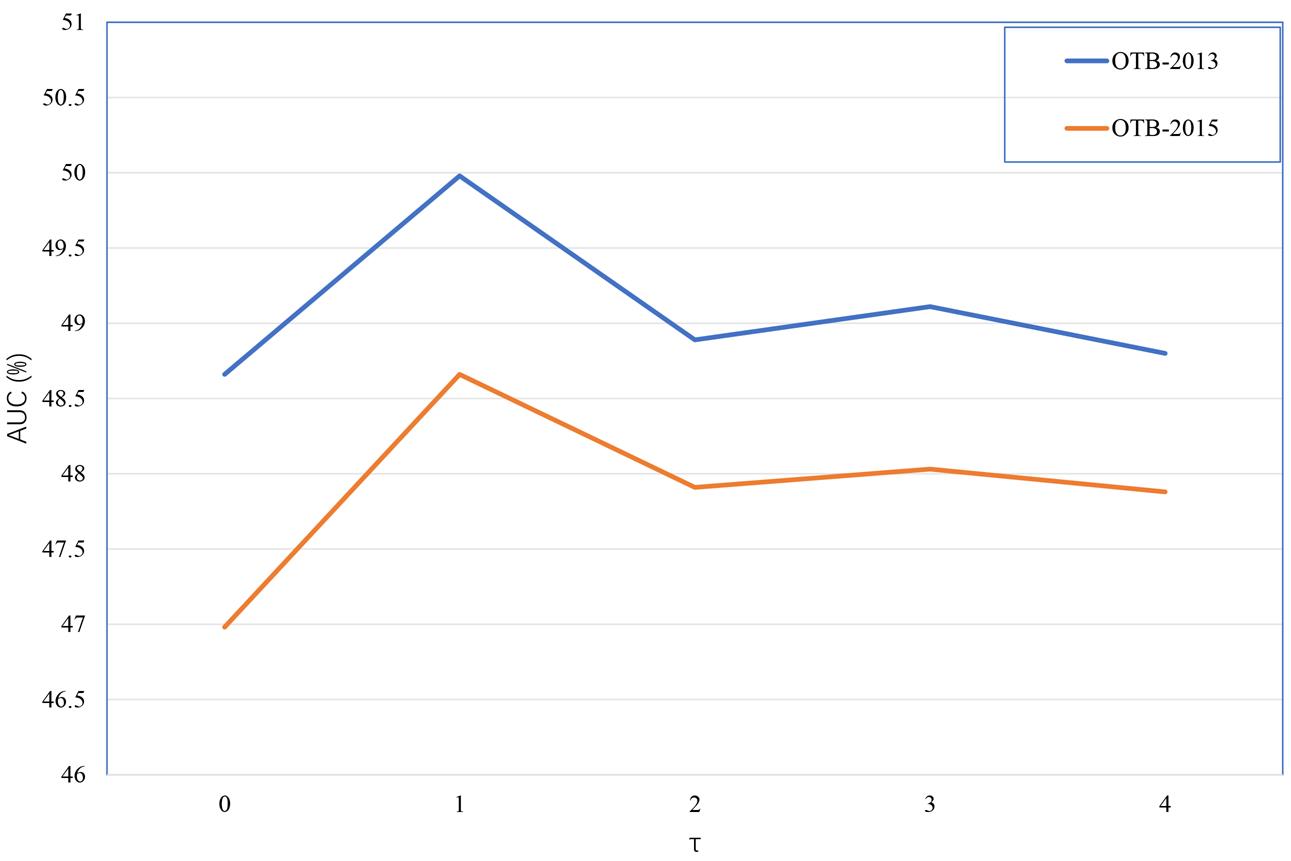}}
	\caption{(a). Results of SiamSNN on different configurations evaluated by latency and the AUC of success plots on OTB-2015. (b). Experimental results for different values of response period threshold on OTB-2013 and OTB-2015.}	
\end{figure}

\subsection{Ablation experiments and hyperparameters settings}
\label{sechp}

We conduct ablation experiments to analyze the effects of the proposed methods and determine the hyperparameters on OTB-2013~\cite{wu2013online} and OTB-2015~\cite{wu2015object}.

To analyze the importance of each proposed component, we report the overall ablation studies in Fig.~\ref{figlatencytao:a}. We gradually add potential similarity estimation (PSE),  potential similarity estimation (PSE), hybrid similarity estimation (HSE), and two-status coding (TS) on SiamSNN converted from SiamFC. As shown in Fig.~\ref{figlatencytao:a}, PSE is the baseline method and the reported scores are tested during 1000 time steps. The AUC score of PSE is 46.36. TSE improves tracking performance from 46.36 to 47.95. HSE brings 0.68 points higher AUC than TSE. Finally, TS obtains a 50.72 AUC score and reduces the latency. This result validates the effectiveness of our proposed methods. SiamSNN+HSE+TS achieves outstanding performance improvement than other configurations.

\textbf{Hybrid spiking similarity estimation.} Fig.~\ref{figlatencytao:b} compares the results for several small values of response period threshold \begin{math}\tau\end{math} on OTB-2013 and OTB-2015. It can be seen that far-distance spikes have poor correlation and increasing the value of \begin{math}\tau\end{math} will bring more computational complexity. Thus, \begin{math}\tau\end{math} is set to 1. 

\textbf{Two-status coding.} As mentioned before in Section~\ref{sechp}, we set \begin{math}\alpha \to+\infty\end{math} ( \begin{math}\alpha = \text{float('Inf')}\end{math} in python code) to prevent neurons from generating spikes during depression status. Refer to the exponential change of voltage threshold in burst coding~\cite{park2019fast}, we set \begin{math}\beta\end{math} to 0.5 to excite neurons spiking during potentiation status. It is inappropriate to set \begin{math}\beta\end{math} to smaller values (e.g., 0.25 and 0.125), which will alter the balanced normalization in Eq.~\ref{eq2},\ref{eqlnorm} and cause worse results. The latency requirements are presented in Fig.~\ref{figlatencytao:a}, we can see that SiamSNN converges rapidly at the beginning and then rises slowly. So in the two-status coding, we choose the state period \begin{math}p\end{math} as 5 to improve the performance in shorter latency. To reach the maximum AUC, SiamSNN+HSE+TS requires approximately 20 time steps.

\subsection{Experimental results}

\begin{table}[]
	\centering
	\caption{Experimental results with other state-of-the-art SNN methods on OTB2013 and OTB2015.}
	\label{tabothers}
	\begin{tabular}{cc|cc|c}
		\toprule
		\begin{tabular}[c]{@{}c@{}}Similarity\\ estimation\end{tabular} & Coding &
		OTB2013 & OTB2015 & Latency \\  \midrule
		\textbf{HSE}    &      rate                        &   48.73           &  46.95           &  600           \\
		\textbf{HSE}    &      phase~\cite{kim2018deep}    &   48.65           &  47.14           &  120           \\
		\textbf{HSE}    &      burst~\cite{park2019fast}   &   48.75           &  47.45           &  100           \\
		\midrule
		\textbf{HSE}    &       \textbf{two-status}        &   \textbf{51.15} &  \textbf{49.31}   &  \textbf{20}  \\
		\bottomrule
	\end{tabular}
\end{table}

\textbf{Comparison with other spike coding methods.}
To the best of our knowledge, SiamSNN is the first deep SNN model for object tracking.
So we compare our coding method with phase~\cite{kim2018deep} and burst coding~\cite{park2019fast} on OTB2013 and OTB2015.
All experiments are performed with HSE due to lacking existed similarity estimation method for SNN tracking. 
As shown in Table~\ref{tabothers}, two-status coding outperforms phase and burst coding, especially on the latency.

Rate coding suffers from an intrinsic flaw about long latency.
For example, 512 time steps are needed to represent the positive integer 512.
To represent 0.006, 6 spikes and 1000 time steps are required.
Phase and burst coding achieve efficient information transmission, which shortens the latency.
Our two-status coding optimizes the temporal distribution of output spike trains.
It enhances the utilization of time steps for similarity estimation, which improves both accuracy and speed.

\textbf{Tracking results.}
Conversion methods of DNN-to-SNN will degrade the accuracy~\cite{kim2020spiking}.
For further evaluation, we implement experiments on VOT-2016 \cite{vot2016}, VOT2018 \cite{vot2018}, and GOT-10k \cite{huang2019got}.
As shown in Table~\ref{tabtrackres}, compared to SiamFC on VOT benchmarks, the accuracy of SiamSNN drops 3.2\% and 1.8\% while the robustness becomes poor, and the degradations of EAO are 2.3\% and 0.7\%.
As for GOT-10k, the AO and SR$_{0.5}$ score of SiamSNN degrades 3.4\% and 2.6\%.
This is the first work that reports the comparable performance of SNNs for object tracking on VOT2016, VOT2018, and GOT-10k datasets.

\begin{table}[]
	\centering
	\caption{Experimental results of performance degradation on VOT2016, VOT2018 and GOT-10k.}
	\label{tabtrackres}
	\begin{tabular}{c|ccc|ccc|cc}
		\toprule
		\multirow{2}{*}{Tracker}                              & \multicolumn{3}{c}{VOT-16} & \multicolumn{3}{c}{VOT-18} & \multicolumn{2}{c}{GOT-10k} \\  
		& A       & R       & EAO    & A       & R       & EAO    & AO          & SR$_{0.5}$      \\ \midrule
		SiamFC    & 0.529   & 0.49    & 0.233  & 0.478   & 0.69    & 0.183  & 0.348       & 0.353  \\ \midrule
		SiamSNN   & 0.497   & 0.63    & 0.210  & 0.460   & 0.86    & 0.176  &  0.314      & 0.327  \\ 
		\bottomrule
	\end{tabular}
	
\end{table}

\textbf{Energy consumption and latency evaluation.}
To investigate the energy-efficient effect of SiamSNN, we evaluate energy consumption of our methods on TrueNorth~\cite{merolla2014million} compared to SiamFC~\cite{bertinetto2016fully}, SiamRPN++~\cite{li2019siamrpn++}, DSTfc~\cite{liu2019teacher} and ECO~\cite{danelljan2017eco}. SiamRPN++ has deeper CNNs and stronger tracking performance that requires additional floating-point operations per second (FLOPS). DSTfc is a small, fast yet accurate tracker by teacher-students knowledge distillation. ECO is a state-of-art correlation-filter based tracker on CPU.

\begin{table}[]
	\centering
	\caption{Comparison of energy consumption of SiamSNN to other trackers.}
	\label{tabenergy}
	\begin{tabular}[width=\linewidth]{ccccc}
		\toprule
		Methods                                                     &  \begin{tabular}[c]{@{}c@{}}FLOPs\\ /SOPS\end{tabular}     & Power (W) & Time (ms)     & Energy (J)    \\ [3pt] \midrule
		SiamFC  & 5.44E+09 & 250      & 11.63        & 2.91          \\[3pt] 
		SiamRPN++  & 1.42E+10 & 250      & 28.57        & 7.14         \\[3pt]
		DSTfc  & -- & 250      & 4.35        & 1.10 \\[3pt]
		ECO  & -- & 120      & 452.49        & 54.3        \\\midrule
		SiamSNN  & \textbf{3.94E+08} & 9.85E-04      & 20 & {\color[HTML]{CB0000} \textbf{1.97E-05}}\\ \bottomrule
		
	\end{tabular}
	
\end{table}

Refer to the reports of these works and~\cite{schuchart2016shift,huang2019got}, we summarize their hardware and speed information as follows. SiamFC(3s) is run on a single NVIDIA GeForce GTX Titan X (250 Watts) and reaches 86 FPS (11.63ms per frame). SiamRPN++ can achieve 35 FPS (28.57ms per frame) on NVIDIA Titan Xp GPU (250 Watts). And DSTfc is run at 230 FPS (4.35ms per frame) on Nvidia GTX 1080ti GPU (250 Watts). ECO is run on an Intel(R) Xeon(R) 2.0GHz CPU (120 Watts) and achieves 2.21 FPS (452.49ms per frame). And TrueNorth measures computation by synaptic operations per second (SOPS) and can deliver 400 billion SOPS per Watt, while FLOPs in modern supercomputers. And the time step is nominally 1ms, set by a global 1kHz clock~\cite{merolla2014million}. We count the operations with the formula in~\cite{rueckauer2017conversion}.

The calculation results of processing one frame are presented in Table~\ref{tabenergy}.
The energy consumption of SiamSNN on TrueNorth is extremely lower than other trackers on GPU or CPU.
Although GPUs are far more advanced computing technology, it is hard to employ DNNs on embedded systems through them.
Neuromorphic chips with higher energy and computational efficiency have a promising development and application.

\section{Conclusion}

In this paper, we propose SiamSNN, the first deep SNN model for object tracking that reaches competitive performance to DNNs with short latency and low precision loss on OTB2013/2015, VOT2016/2018, and GOT-10k.
Consequently, we propose an optimized hybrid spiking similarity estimation method and a two-status coding scheme for taking full advantage of temporal information in spike trains, which achieves real-time on TrueNorth.
We aim at studying the spiking representation of SiamRPN and SiamRPN++ in the subsequent research.
We believe that our methods can be applied in more spiking Siamese networks for energy-efficient tracking or other similarity estimation problems.

%
%
%
 \bibliographystyle{splncs04}
 \bibliography{mybibliography}

\end{document}